\documentclass{article}

\usepackage[nonatbib,final]{neurips_2020}

\usepackage[utf8]{inputenc} % allow utf-8 input
\usepackage[T1]{fontenc}    % use 8-bit T1 fonts
\usepackage{hyperref}       % hyperlinks
\usepackage{url}            % simple URL typesetting
\usepackage{booktabs}       % professional-quality tables
\usepackage{amsfonts}       % blackboard math symbols
\usepackage{nicefrac}       % compact symbols for 1/2, etc.
\usepackage{microtype}      % microtypography

\usepackage{amsmath}     
\usepackage{algorithm}
\usepackage[noend]{algpseudocode}
\usepackage{ctable}
\usepackage{multirow}
\usepackage{etoolbox}  % for space between pseudocode
\usepackage{tabularx}  
\usepackage{arydshln}
\usepackage{subcaption}
\usepackage{kotex}

\newcommand{\prgname}{FOCI}

\AtBeginEnvironment{algorithm}{\linespread{1.1}\selectfont}

\newcommand{\Lagr}{\mathcal{L}}

\newcommand{\ie}{\textit{i}.\textit{e}.}

\title{ Which Strategies Matter for Noisy Label Classification? Insight into Loss and Uncertainty }

\author{Wonyoung Shin\textsuperscript{\rm 1}, 
Jung-Woo Ha\textsuperscript{\rm 2},
Shengzhe Li\textsuperscript{\rm 1},
Yongwoo Cho\textsuperscript{\rm 1},
Hoyean Song\textsuperscript{\rm 2},
Sunyoung Kwon\textsuperscript{\rm 2}\thanks{∗Correspondence should be addressed to Sunyoung Kwon.}
\\ 
\textsuperscript{\rm 1}Naver Shopping, NAVER Corp.\\ 
\textsuperscript{\rm 2}Clova AI Research, NAVER Corp.\\ 
{\tt\small \{wonyoung.shin,jungwoo.ha,s.li,yongwoo.cho,chris.ai,sunny.kwon\}@navercorp.com 
}}

\newcounter{alphasect}
\def\alphainsection{0}

\let\oldsection=\section
\def\section{%
  \ifnum\alphainsection=1%
    \addtocounter{alphasect}{1}
  \fi%
\oldsection}%

\renewcommand\thesection{%
  \ifnum\alphainsection=1% 
    \Alph{alphasect}
  \else%
    \arabic{section}
  \fi%
}%

\newenvironment{alphasection}{%
  \ifnum\alphainsection=1%
    \errhelp={Let other blocks end at the beginning of the next block.}
    \errmessage{Nested Alpha section not allowed}
  \fi%
  \setcounter{alphasect}{0}
  \def\alphainsection{1}
}{%
  \setcounter{alphasect}{0}
  \def\alphainsection{0}
}%

\begin{document}

% \vskip 0.3in

\maketitle

\begin{abstract}
Label noise is a critical factor that degrades the generalization performance of deep neural networks, thus leading to severe issues in real-world problems. Existing studies have employed strategies based on either loss or uncertainty to address noisy labels, and ironically some strategies contradict each other: emphasizing or discarding uncertain samples or concentrating on high or low loss samples. To elucidate how opposing strategies can enhance model performance and offer insights into training with noisy labels, we present analytical results on how loss and uncertainty values of samples change throughout the training process. From the in-depth analysis, we design a new robust training method that emphasizes clean and informative samples, while minimizing the influence of noise using both loss and uncertainty. We demonstrate the effectiveness of our method with extensive experiments on synthetic and real-world datasets for various deep learning models. The results show that our method significantly outperforms other state-of-the-art methods and can be used generally regardless of neural network architectures.

\end{abstract}

\section{Introduction}

Recent advances in deep learning have significantly improved performance in numerous tasks due to large quantities of human-annotated data. While standard large-scale benchmark datasets used for deep learning research such as ImageNet~\cite{deng2009imagenet} are generally clean and error-free, most real-world data contain noisy labels, which refer to observed labels that are incorrect~\cite{frenay2013classification}. Because obtaining reliably labeled data is expensive, labor-intensive and time-consuming, label noise is common and inevitable in most real-world datasets. 

The ubiquity of noise is all the more a critical issue for it is known that learning with noisy labels severely degrades model performance. As reported by Zhang et al.~\cite{zhang2016understanding}, deep neural networks are capable of fitting random noisy labels. If even a small portion of noisy labels exists within the training data, deep learning models can eventually memorize the wrongly given labels, thus deteriorating performance. It is, therefore, necessary to design methods that are robust to label noise such that negative consequences are minimized.

\begin{figure}[t]
\centering
\includegraphics[width=0.55\textwidth]{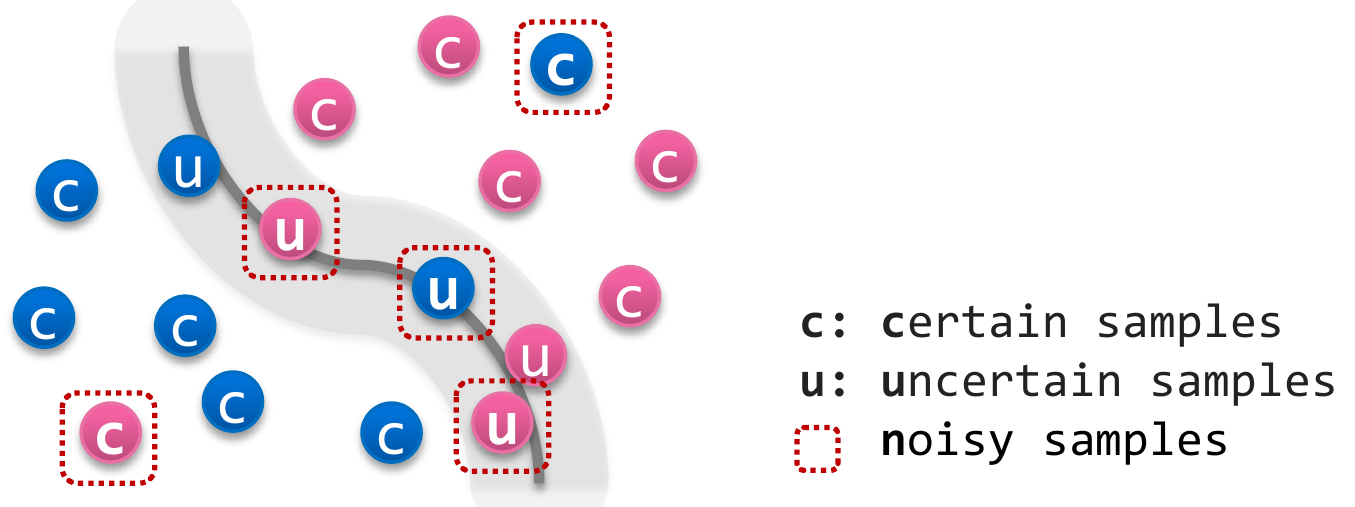} 
\caption{Label noise and sample uncertainty. Uncertain samples are located near the decision boundary and cause predictions to change constantly. Samples with high certainty are far from the decision boundary and lead to consistent predictions. Noisy samples can be close to or far from the decision boundary and are easily identifiable when they are distant from the decision boundary. Each color represents a different class. } 
\vspace{-1.1em}
\label{fig:noise}
\end{figure}

One approach for dealing with noisy labels is to focus on samples according to their uncertainty during the training phase (see \figurename~\ref{fig:noise}). Some methods emphasize uncertain samples, the predictions of which are inconsistent during training~\cite{chang2017active,malach2017decoupling}. As reported in previous studies on active learning, these uncertain samples are informative and require more training than other samples~\cite{schein2007active, joshi2009multi}, while samples that are well-trained and have consistent predictions have less information in improving the model. Performance can consequently be boosted by preferring uncertain samples which are near the class decision boundary. On the other hand, some methods reduce the importance or exclude uncertain samples so that only highly certain samples remain in the training data~\cite{song2019selfie}. Although the impact of informative samples is minimized, it can produce a coherent model and be a safer way of training.

% Samples with high certainty are less informative than uncertain samples, because they have already been trained thoroughly.

Another way to address noisy labels is by managing samples depending on their loss. Loss can signify the difficulty and the confidence of predictions, so giving precedence to samples with low loss or samples with high loss can work well depending on the amount of noise in the data or the complexity of the problem~\cite{hinton2007recognize,loshchilov2015online}. Difficult samples are known to accelerate training, especially for datasets with a small amount of noise~\cite{malisiewicz2011ensemble}. For this reason, there have been studies that increase the weights of high loss samples so that the network focuses on difficult samples~\cite{shrivastava2016training}. More recently, however, researchers have somewhat ironically taken the opposite approach by emphasizing easy samples~\cite{han2018co,huang2019o2u,chen2019understanding}. Because easy samples are likely to be clean, favoring low loss samples has been proven to enhance performance, especially when solving a difficult task such as training with severe label noise~\cite{chang2017active}.

% for heavier noise. easy samples are effective 

% ~\cite{huang2019o2u, chen2019understanding, song2019selfie}

\begin{table}[t]
    \centering
    \caption{ Details of datasets used in the experiments. }
    \label{tab:data}
    \footnotesize 
    \begin{tabular}{ccccc}
        \toprule
         Dataset & Train \# & Test \# & Class \# & Image size \\
        %dataset & \# of train & \# of test & \# of classes & image size \\
        \midrule
        CIFAR-10            & 50K                 & 10K                & 10            & 32x32      \\ 
        CIFAR-100           & 50K                 & 10K                & 100           & 32x32      \\ 
        Tiny ImageNet$^\ast$     & 100K                & 10K                & 200           & 64x64      \\ 
        Clothing1M     & 1M noisy                & 10K                & 14           & 256x256      \\ 
        \bottomrule 
        \multicolumn{5}{l} {%
        \begin{minipage}{6.5cm}%
            \smallskip
            $^\ast$a subset of ImageNet~\cite{deng2009imagenet} 
        \end{minipage}%
        }
    \end{tabular}
    \vspace{-1.2em}
\end{table}

In the literature, it is shown that contrasting strategies effectively diminish the effect of noisy samples, leading to improved performance over the baseline. Motivated to understand how all approaches can enhance accuracy, we analyze the changing loss and uncertainty of samples in the course of training for CIFAR-10, CIFAR-100, and Tiny ImageNet with different noise types. Data show that symmetric noise is easy to identify using either loss or uncertainty, whereas asymmetric noise is challenging to distinguish from clean samples, indicating the need for an efficient alternative method.

%with the remaining samples, emphasize those that are likely to be clean and informative. 

Inspired by the finding that only a minority of samples with low loss and high uncertainty have noisy labels, we propose \prgname{} (\textit{Focus On Clean and Informative samples}), a novel robust training method. Our key idea is to emphasize the samples that are likely to be clean and informative. \prgname{} prioritizes samples with low loss and high uncertainty and minimizes the impact of samples of high loss since they are very likely to be noisy. To validate our method, we conducted extensive experiments on CIFAR-10, CIFAR-100, and Tiny ImageNet with diverse noise types from 40\% to 70\% of noise levels, as well as on a real-world dataset Clothing1M. Moreover, we observed the performance of training with various deep learning models to check generalizability. Our empirical analysis demonstrates the enhanced robustness of \prgname{} on noisy datasets, and its generalizability to any network architecture, making \prgname{} a useful addition to real-world deep learning pipelines.

The contribution of this paper is three-fold. (1) We identify insights on how loss and uncertainty affect noisy label classification via an in-depth analysis. 
(2) Inspired by these insights, we design a novel lightweight method that robustly learns by focusing on clean and informative samples from data with various conditions and types of noisy labels without any additional clean data. 
(3) With thorough experiments, we demonstrate our~\prgname{}'s robustness to label noise that substantially outperforms state-of-the-art methods on a real-world dataset and three benchmark datasets injected with diverse synthetic noise.

\begin{figure}[t]
\centering
\begin{minipage}[t]{\textwidth}
\centering  
    \includegraphics[width=\textwidth]{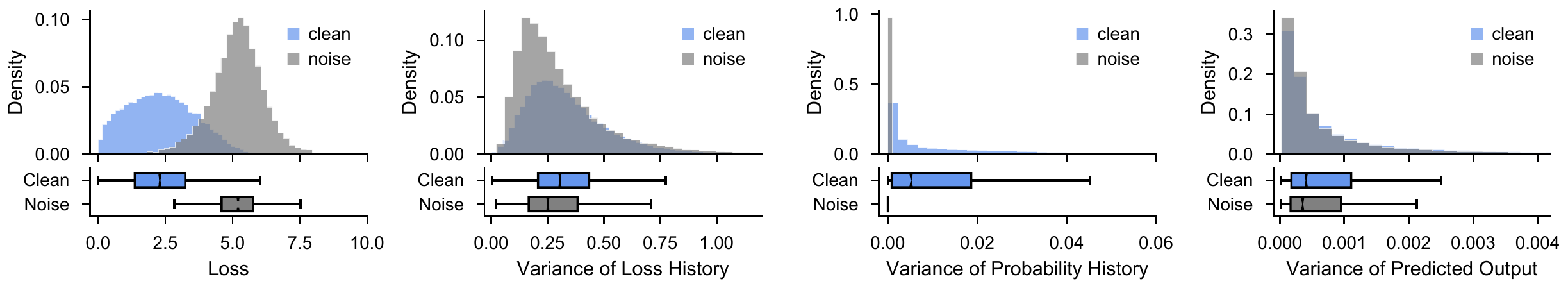}
    \subcaption{Symmetric noise.}
    \label{fig:sample-weights-10}
\end{minipage}%
\vspace{2pt}
\begin{minipage}[t]{\textwidth}
\centering
    \includegraphics[width=\textwidth]{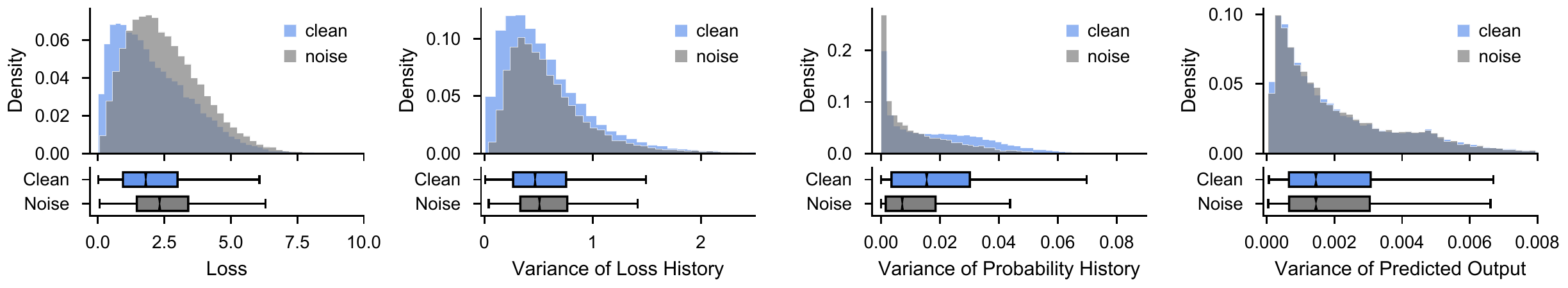}
    \subcaption{Asymmetric noise.}
    \label{fig:sample-weights-100}
\end{minipage}%
\vspace{-2pt}
\caption{Normalized distribution of loss and uncertainty on CIFAR-100 with 40\% noise at epoch 50. }
\label{fig:loss_var_e50}
\vspace{-0.2em}
\end{figure}

\begin{figure}[t]
\centering
\includegraphics[width=0.85\textwidth]{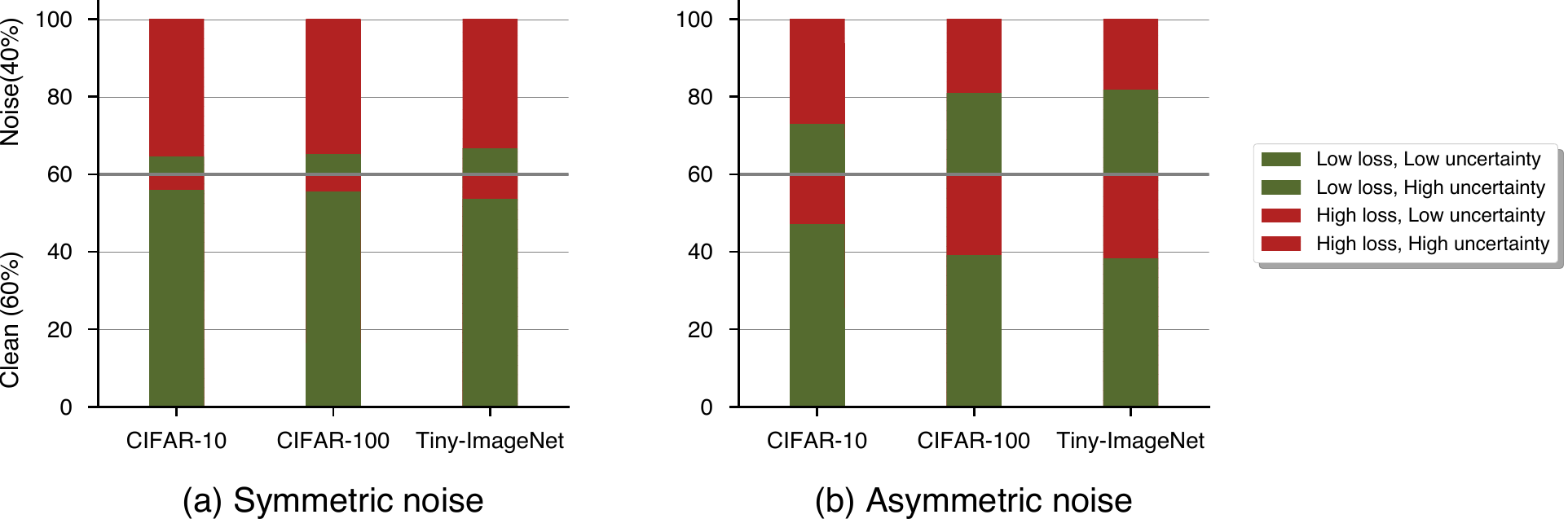} 
\caption{Combinations of loss and uncertainty 
%for CIFAR-10, CIFAR-100, and Tiny ImageNet 
with 40\% noise rate at epoch 50. Clean (60\%) and noisy (40\%) samples are divided vertically. The green and red colors represent low and high loss samples, respectively. The solid and stripe patterns represent low and high variation samples, respectively. } 
\vspace{-1em}
\label{fig:loss_variance_datasets_50}
\end{figure}

\section{Loss and Uncertainty in Noisy Datasets} \label{loss_var_analysis}

We explore how loss and uncertainty differ for various label noise by training DenseNet (L=25, k=12, momentum optimizer) on three benchmark datasets, listed in \tablename~\ref{tab:data}: CIFAR-10, CIFAR-100, and Tiny ImageNet. These datasets are commonly used to evaluate noisy labels~\cite{goldberger2016training,reed2014training}. 
We artificially corrupted the data by following typical protocols~\cite{han2018co, van2015learning}. In accordance with prior studies, for $k$ classes, noise is given by swapping true labels for other class labels with some constant probability, namely, noise rate $\tau$~\cite{han2018co}. In this experiment, we set $\tau=0.4$. While labels are swapped between two classes for asymmetric noise, labels are swapped to classes other than the true class label with probability of $\frac{\tau}{k-1}$ for symmetric noise~\cite{jiang2017mentornet, wang2018iterative}. Each noise type is described in Appendix A.

To analyze uncertainty in noisy datasets, we explore various representations of uncertainty. Uncertainty can be quantified by the variance of loss or predicted probabilities for a given class in a $q$-sized history~\cite{chang2017active}. Another definition is by the variance of the predicted probabilities over all the classes at one step~\cite{konyushkova2017learning,yang2018variance}. The normalized distributions of each definition where $q=15$ are displayed in \figurename~\ref{fig:loss_var_e50}. The difference of distributions between noisy and clean samples is more pronounced for variance based on the history of predicted probabilities, so we use this definition for uncertainty.

As can be seen from \figurename~\ref{fig:loss_variance_datasets_50}, we check samples based on their loss and uncertainty (prediction variance) by dividing them into four groups. Samples are split into low loss and high loss with a ratio of $1-\tau : \tau$ as suggested by Han et al.~\cite{han2018co}. The same applies for uncertainty in which the ratio of low uncertainty and high uncertainty samples is $1-\tau : \tau$. The proportions concerning loss and uncertainty did not change much during training, so we only display the results at epoch 50 due to lack of space. The detailed results can be found in Appendix B.

\subsection{Analysis of symmetric noise}
%\subsection{Results on symmetric noise}
As shown in Figures~\ref{fig:loss_var_e50}a and~\ref{fig:loss_variance_datasets_50}a, clean and noisy samples have distinct characteristics for symmetric noise, seemingly due to the easiness of symmetric noise and deep neural networks' capability of generalizing on data with symmetric noise~\cite{rolnick2017deep}. The majority of clean samples have lower loss and relatively higher uncertainty than noisy samples, providing evidence for enhanced accuracy of approaches which emphasize high uncertainty. In contrast to clean samples, most noisy samples have higher loss and lower uncertainty. The loss of noisy samples tends to be higher than those of clean samples because predictions are different from the given labels, and the uncertainty of noisy samples is close to 0 because the predicted probabilities for the given noisy labels maintain a very small value. Approaches that emphasize easy samples (low loss) or uncertain samples can thus benefit from this fact. These findings not only support our idea of emphasizing low loss and high uncertainty samples, but also confirm that symmetric noise is an easy problem to be solved and less practical as stated in prior works~\cite{han2018co,ren2018learning}.

\subsection{Analysis of asymmetric noise}
%\subsection{Results on asymmetric noise}
Inspection of \figurename~\ref{fig:loss_variance_datasets_50}b indicates that noisy samples can have high or low loss and uncertainty, thus justifying the enhanced performance of strategies that contradicted each other. However, according to \figurename~\ref{fig:loss_var_e50}b, it is not effective to distinguish clean and noisy samples solely based on loss or uncertainty; the loss of clean and noisy samples are alike, and the difference between the uncertainty of clean and noisy samples is very subtle. Taking both loss and uncertainty into consideration seems more effective and plausible when training data with asymmetric noise, which is problematic and similar to real-world noise~\cite{ren2018learning, yi2019probabilistic}.

\renewcommand{\algorithmiccomment}[1]{\hfill \(\triangleright\)~#1}
\begin{algorithm*}[t]
\caption{ \prgname{} Algorithm }\label{pseudocode}
\textbf{Input:} mini-batch $\mathcal{D}_b$ from dataset $\mathcal{D}$
\begin{algorithmic}[1]
% \State $\mathbf{C} \gets \emptyset$
\For {$t \gets 1$ to $T$}
        \If {$\textit{t} \leq \gamma$}
        \algorithmiccomment {during warm-up ($\gamma$) phase}
            \State $\theta \gets \theta - \alpha \nabla \frac{1}{N_b} \sum \Lagr(x,y;\theta)$
            \algorithmiccomment {parameter update by $\Lagr$ from $\mathcal{D}_b$}
        \Else
        \algorithmiccomment {after warm-up, \textbf{\prgname{}} phase}
%            \State $\mathcal{H} \gets \tau\times100 \%$ of high loss samples in $\mathcal{D}_b$
            % \State $\mathcal{R} \gets (\mathrm{E} > \epsilon \cap x \in \mathcal{H}) \cup (\mathrm{E} \leq \epsilon \cap \hat{y} \neq y )$
            % \algorithmiccomment {sample removal ($\mathcal{R}$)}
            \State $W \gets {\mathrm{normalize}}(\sqrt{P_t(y|x) \cdot \mathrm{Var}(P_{t-q+1:t}(y|x))}) $
            \algorithmiccomment {sample weighting ($W$)}
            \State $\theta \gets \theta - \alpha \nabla \frac{1}{N_b} \sum  W(x,q)\Lagr(x,y;\theta)$
            \algorithmiccomment {parameter update by $W$$\Lagr$}

            %$\theta \gets \theta - \alpha \nabla (\frac{1}{|\mathcal{E} \cup \mathcal{C}|} \sum_{x \in \mathcal{E} \cup \mathcal{C}} W(x,q)\Lagr(x,y;\theta))$
            % \State $\mathbf{C} \gets \mathbf{C} \cup \mathcal{C}$
        \EndIf
\EndFor
\end{algorithmic}
\end{algorithm*}

\section{Method}  
\subsection{Overview}
To validate findings from our analysis, we design a novel lightweight method \prgname{} that aims at focusing on clean and informative samples. The overall procedure of our method is summarized in Algorithm~\ref{pseudocode}. 

Let the training set be $\mathcal{D} = \{(x,y)\}$ of size $N$, and the dataset for a mini-batch be $\mathcal{D}_b$ of size $N_b$. 
When training a network parameter $\theta$ in the warm-up phase with learning rate $\alpha$ (Lines 2-3), updating parameters can be formulated as:
\begin{equation} \label{eq:standard_loss}
\theta \gets \theta - \alpha \nabla (\frac{1}{N_b} \sum_{x \in \mathcal{D}_b} \Lagr(x,y;\theta)),
\end{equation}

The algorithm starts by updating the network in the conventional way stated above. This is because deep neural networks can learn simple and common patterns, even with the presence of noisy labels during the early warm-up phase~\cite{jiang2017mentornet,arpit2017closer}. However, since real-world datasets are bound to have noise, our method pursues robust training after the warm-up phase (Lines 4-6) by reweighting samples so clean and informative samples are emphasized and the impact of noisy samples are minimized:
\begin{equation} \label{eq:our_loss}
\theta \gets \theta - \alpha \nabla (\frac{1}{N_b} \sum_{x \in D_b} W(x,q)\Lagr(x,y;\theta)),
\end{equation}
where $W(x,q)$ is the reweighting function.

\setlength{\tabcolsep}{6pt}
\setlength{\floatsep}{4pt}
\setlength{\textfloatsep}{12pt}

\ctable[
%  	captionskip = -10pt,
    caption = {Classification accuracy (\%) on benchmark datasets with $40\%$ noise. Asymmetric and symmetric noise are denoted by A and S respectively.  },
    label = tab:all,
    doinside = \footnotesize,
    pos=t
]{cl|ccccc}{
%\tnote[$\ast$] {We experimented SELFIE with 1st run to equalize the total number of epochs with others}
}{
\toprule

\multicolumn{1}{l}{}              &               & Default      & Active Bias   & Coteaching   & SELFIE                                 & \prgname                \\
\midrule
\multirow{3}{*}{Asymmetric noise} & CIFAR-10      & 71.8$\pm$1.5 & 78.0$\pm$1.5 & 83.7$\pm$1.4 & 84.9$\pm$0.1                           & \textbf{86.2$\pm$0.4} \\
                                  & CIFAR-100     & 45.3$\pm$1.4 & 50.3$\pm$0.6 & 47.3$\pm$1.4 & 52.8$\pm$0.5                           & \textbf{59.5$\pm$0.9} \\
                                  & Tiny ImageNet & 30.8$\pm$0.1 & 33.2$\pm$0.9 & 30.3$\pm$0.5 & 36.1$\pm$0.3                           & \textbf{37.6$\pm$0.9} \\ \midrule

 \multirow{3}{*}{\begin{tabular}[c]{@{}c@{}}Mixed noise \\ (A-$30$, S-$10$)\end{tabular}} & CIFAR-10      & 79.6$\pm$0.8 & 84.9$\pm$0.5 & 80.6$\pm$1.7 & 84.9$\pm$0.9 & \textbf{85.7$\pm$0.4} \\
                                                                                         & CIFAR-100     & 49.9$\pm$0.7 & 56.2$\pm$0.5 & 50.9$\pm$0.8 & 58.5$\pm$0.3 & \textbf{61.5$\pm$0.5} \\
                                                                                         & Tiny ImageNet & 35.0$\pm$0.8 & 36.2$\pm$0.4 & 34.0$\pm$0.6 & 39.0$\pm$0.6 & \textbf{39.7$\pm$0.7} \\  \midrule
\multirow{3}{*}{\begin{tabular}[c]{@{}c@{}}Mixed noise \\ (A-$20$, S-$20$)\end{tabular}} & CIFAR-10      & 81.2$\pm$0.8 & 84.8$\pm$0.3 & 82.1$\pm$0.3 & 84.8$\pm$0.4 & \textbf{86.1$\pm$0.9} \\
                                                                                         & CIFAR-100     & 53.0$\pm$0.4 & 58.2$\pm$1.1 & 54.2$\pm$1.0 & 59.1$\pm$0.5 & \textbf{60.6$\pm$0.5} \\
                                                                                         & Tiny ImageNet & 36.8$\pm$0.9 & 37.6$\pm$0.6 & 37.1$\pm$1.5 & 37.9$\pm$0.2 & \textbf{37.9$\pm$0.2} \\  
                                                                                         \midrule
                           Nearest noise       & CIFAR-100     & 45.8$\pm$0.8 & 54.8$\pm$0.7 & 55.9$\pm$0.8 & 57.8$\pm$0.3                           & \textbf{57.9$\pm$0.5} \\
% \multirow{2}{*}{Nearest noise}    & CIFAR-10      & 67.9$\pm$1.1 & 74.7$\pm$1.0 & 83.3$\pm$0.5 & \textbf{83.5$\pm$0.4} & 83.4$\pm$0.0                           \\
                                %   & CIFAR-100     & 45.8$\pm$0.8 & 54.8$\pm$0.7 & 55.9$\pm$0.8 & 57.8$\pm$0.3                           & \textbf{57.9$\pm$0.5} \\
 
\bottomrule
}

\subsection{Sample weighting}

Our aim is to ensure that clean and informative samples contribute more to training, so we place more importance on samples with low loss and high uncertainty. Because the loss value and the predicted softmax probability are inversely proportional to each other, emphasizing samples with a high predicted probability of the given label would more or less be identical to focusing on samples with low loss. As a result, to favor clean samples, we compute $W(x,q)$ using $P_t(y|x)$, the predicted probability of the given label, and $\mathrm{Var}(P_{t-q+1:t}(y|x))$, the variance of predicted probabilities in the history queue for epochs from $t-q+1$ to $t$. 
\begin{equation} \label{eq:m}
W(x,q) = \mathrm{normalize}\sqrt{P_t(y|x) \cdot \mathrm{Var}(P_{t-q+1:t}(y|x))}
%m(x,q) = \sqrt{P_t(y|x) \cdot \mathrm{Var}(P_{t-q+1:t}(y|x))}
\end{equation}
%$m(x,q)$ 
The weights are subsequently standardized (\ie, mean is 0 and standard deviation is 1), and bounded with the sigmoid function to give a clipping effect, and are further divided by a normalizing factor to have unit mean. These normalized sample weights $W(x,q)$ are multiplied to the loss function, allowing cleaner samples to contribute more when updating the network (Line 6). 

We also reduce the impact of samples that are likely to be noisy using methods partially based on \textit{SELFIE}. We screen samples with inconsistent predictions and high loss or samples with consistent predictions but the predicted label disagrees from the given label, and set their weights to zero.

Inconsistency is represented by a normalized information entropy of label frequency \ie, $-\frac{1}{\log(q)}\sum_{\hat{y}=1}^{k}F(\hat{y})\log F(\hat{y})$, where $F$ denotes the frequency proportion of label $\hat{y}$ in the $q$ sized prediction history, and $\hat{y}$ denotes each predicted class label of $k$ classes. %, and $\delta=$, a normalization term that scales values to [0,1]. 
Samples that have inconsistency values higher than a certain threshold $\epsilon \in [0,1]$ are considered noisy because their predicted classes have changed constantly during training. To identify high loss samples, we adopt the widely used loss-based separation method. Loss values ranked in the top $\tau \times 100\%$ within the minibatch are classified as high loss. The noise rate $\tau$ can be estimated through cross-validation if unknown~\cite{li2017learning,liu2015classification}.

\section{Experiments}

\noindent\emph{\textbf{Datasets and label corruption schemes}.}  To validate the effectiveness of our method, we perform an image classification task on three benchmark datasets: CIFAR-10\footnotemark, CIFAR-100\footnotemark[\value{footnote}], \footnotetext{https://www.cs.toronto.edu/\textasciitilde{}kriz/cifar.html} and Tiny ImageNet\footnote{https://www.kaggle.com/c/tiny-imagenet} and a real-world dataset Clothing1M~\cite{xiao2015learning} (see \tablename~\ref{tab:data}). For the benchmark datasets, we use four label corruption schemes: symmetric noise, asymmetric noise, mixed noise, and nearest label transfer. In this work, we are concerned with scenarios of abundant data with very poor but realistic label quality. Because labelers make mistakes within very few and similar classes~\cite{han2018co,ren2018learning, yi2019probabilistic}, asymmetric noise is injected to these datasets with varying noise rates $\tau \in \{0.1, 0.2, 0.3, 0.4\}$, and symmetric noise is mixed with asymmetric noise. To simulate confusions between visually similar classes, we also employ nearest label transfer~\cite{seo2019combinatorial}, in which labels are swapped according to a confusion matrix of a pretrained network. All the noise types are detailed in Appendix A.

\noindent\emph{\textbf{Experimental settings.}}
We use two different schemes for the learning rate policy and number of epochs depending on the type of noise that is used. For symmetric noise, we follow the experimental settings of Arazo et al.~\cite{arazo2019unsupervised} and train PreAct ResNet-18 using SGD with a momentum of 0.9, weight decay of $10^{-4}$, and a batch size of 128 for 300 epochs. The initial learning rate is 0.1 and reduced by a factor of 10 at epoch 100 and 250. We set $\epsilon=0.1$, $q=25$, $\gamma=250$ for CIFAR-10 and $\gamma=100$ for CIFAR-100. Data preprocessing and augmentation is also applied, including mean subtraction, horizontal random flip, 32x32 random crops after padding with 4 pixels on each side, and mixup augmentation~\cite{zhang2018mixup}. We report the best classification accuracy (\ie, the percentage of correct predictions out of the entire test dataset) across epochs following prior works~\cite{seo2019combinatorial, arazo2019unsupervised}. For other types of noise, we trained DenseNet (L=25, k=12) for 100 epochs using SGD with a momentum of 0.9 in line with experiments conducted by Huang et al.~\cite{huang2017densely}. The initial learning rate is 0.1 and divided by 5 at epoch 50 and 75. We use batch size of 128, $\epsilon=0.1$, $q=15$, and $\gamma=25$. Each image is scaled to have zero mean and unit variance. We measure performance by the mean of last classification accuracies over three runs for it is common to measure the robustness of noisy labels with the test error at the end of training~\cite{malach2017decoupling, song2019selfie}. We also compute the label precision by the fraction of true clean samples among all the samples selected for training or samples that have non-zero weights. All of the experiments were executed using NAVER Smart Machine Learning (NSML) platform~\cite{kim2018nsml,sung2017nsml}.

\begin{figure}[t]
\centering
\includegraphics[width=0.95\textwidth]{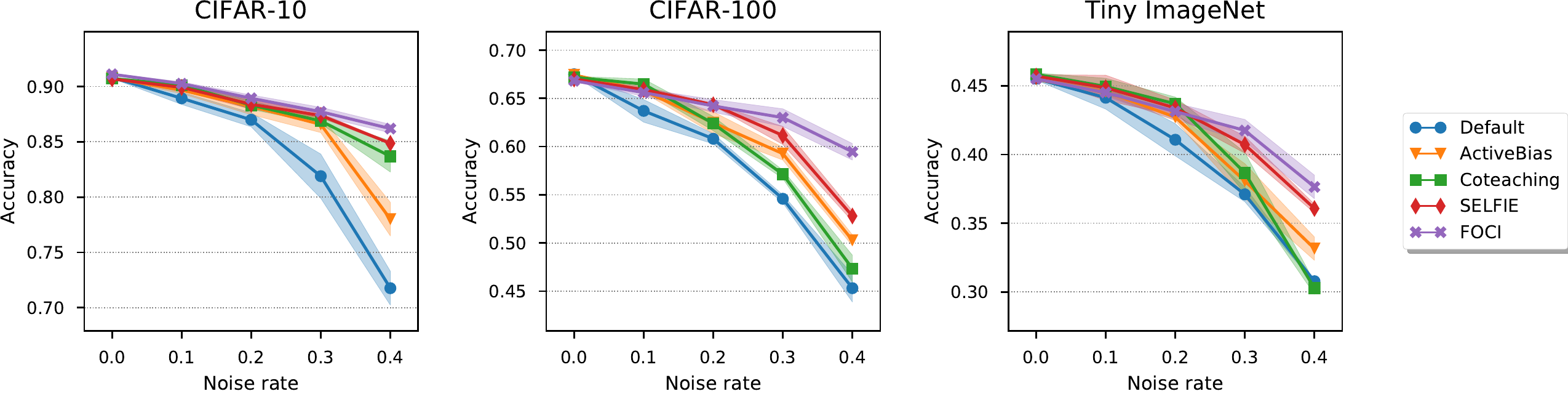} 
\caption{Classification accuracy comparison on three benchmark datasets with varying rates of \textbf{asymmetric noise}. The shaded area represents the standard deviation of three repeated experiments.
\label{fig:asymmetric_noise_rate}}
\vspace{0.2em}
\end{figure}

\begin{figure}[t]
\centering
\includegraphics[width=0.95\textwidth]{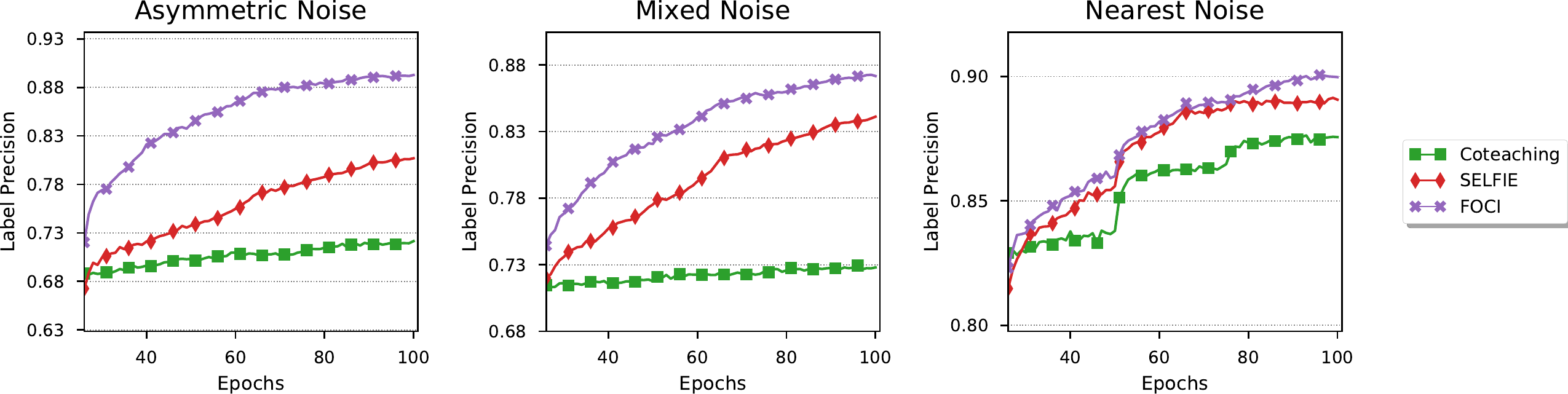} 
\caption{Label precision comparison on CIFAR-100 with $40\%$ noise after the warmup phase (epoch $25$). Mixed noise comprises $30\%$ asymmetric noise and $10\%$ symmetric noise. \textit{Default} and \textit{Active Bias} do not distinguish clean samples, so they are not included for comparison.
}
\label{fig:precision}
\end{figure}

%%%%%%%%%%%%%%%%%%%%%%%%%%%%%%%%%%%%%%%%%%%%%%%%%%%%%%%%%%%%%%%%%%%%%%%%%%%%%%%%%%%%%%%%%%%%%%%%%%%%%%%%%%%%%%%%%%%%%%%%%%%%%%%%%%
\subsection{Performance Comparison}

\noindent\emph{\textbf{Baselines.}}
We compare \prgname{} with a baseline algorithm (denoted by \textit{Default}), which trains noisy data without any strategies, %method, 
an uncertainty-based approach \textit{Active Bias}, loss-based approach \textit{Coteaching}, and a hybrid approach \textit{SELFIE}. \textit{Active Bias}~\cite{chang2017active} emphasizes uncertain samples with high prediction variances. \textit{Coteaching}~\cite{han2018co} uses two networks that feed each other low loss samples. \textit{SELFIE}~\cite{song2019selfie} selects low loss samples and relabels samples with high certainty. It is fair to compare algorithms with the same number of epochs, so we did not restart \textit{SELFIE}, which caused different results from the paper. \prgname{} can handle label noise with only a noisy train dataset, so we did not compare methods that require an additional clean dataset~\cite{shu2019meta}.

%% Asymmetric noise
\noindent\emph{\textbf{Asymmetric noise.}}
\figurename~\ref{fig:asymmetric_noise_rate} displays the test accuracies of \prgname{} along with other baseline methods for varying rates of asymmetric noise. It appears that the performance of \textit{Default} degrades drastically as the noise rate increases. Although other methods achieve higher accuracies than those of \textit{Default}, our method outperforms all other baselines with significant margins for each dataset and noise rate. Moreover, as can be seen from \tablename~\ref{tab:all}, there is a remarkable improvement in performance for CIFAR-100 where the accuracy differs with the second-best algorithm by 7\%. \figurename~\ref{fig:precision} also shows that our method is effective at detecting and filtering out noise even for the difficult scenario of asymmetric noise.

\setlength{\tabcolsep}{6pt}
\setlength{\floatsep}{4pt}
\setlength{\textfloatsep}{12pt}

\ctable[
% 	captionskip = -5pt,
    %caption = {Classification accuracy(\%) on \textbf{mixed noise} using DenseNet (L=25, k=12).},
    pos=t,
    caption = {Classification accuracy (\%) on CIFAR-100 with high-level noise. Asymmetric and symmetric noise are denoted by A and S respectively.},
    label = tab:largenoise,
    doinside = \footnotesize
]{cc|ccccc}{
}{
\toprule

\multicolumn{1}{l}{} &                   & Default      & Active Bias   & Coteaching   & SELFIE       & \prgname{}            \\

\midrule

Mixed noise          & 50\% (A-40, S-10) & 38.0$\pm$1.2 & 41.2$\pm$1.0 & 37.3$\pm$0.9 & 42.1$\pm$2.7 & \textbf{48.8$\pm$2.1} \\
Mixed noise          & 60\% (A-30, S-30) & 35.9$\pm$0.4 & 40.8$\pm$1.5 & 37.0$\pm$0.7 & 43.8$\pm$0.4 & \textbf{48.5$\pm$1.4} \\
Mixed noise          & 70\% (A-20, S-50) & 32.9$\pm$0.7 & 35.8$\pm$0.5 & 32.2$\pm$0.2 & 41.5$\pm$0.9 & \textbf{42.0$\pm$2.2} \\
Nearest noise        & 60\%              & 36.3$\pm$0.9 & 43.2$\pm$0.1 & 44.0$\pm$1.6 & 46.6$\pm$1.1 & \textbf{47.1$\pm$0.9} \\

\bottomrule
}

\ctable[
% 	captionskip = -5pt,
    pos=t,
    caption = {Classification accuracy (\%) of various architectures on CIFAR-100 with $40\%$ asymmetric noise.},
    label = tab:modelcomparison,
    doinside = \footnotesize,
]{l|ccccc}{
}{
\toprule
            & Default      & Active Bias   & Coteaching   & SELFIE       & \prgname{}            \\
\midrule
DenseNet    & 45.3$\pm$1.4 & 50.3$\pm$0.6 & 47.3$\pm$1.4 & 52.8$\pm$0.5 & \textbf{59.5$\pm$0.9} \\
VGG-19      & 35.4$\pm$1.9 & 31.4$\pm$0.7 & 35.6$\pm$0.5 & 39.3$\pm$0.3 & \textbf{43.1$\pm$0.6} \\
ResNet50    & 29.8$\pm$0.2 & 28.2$\pm$0.5 & 32.6$\pm$0.2 & 32.8$\pm$0.2 & \textbf{35.9$\pm$0.4} \\
MobileNetV2 & 32.9$\pm$0.6 & 38.1$\pm$0.6 & 31.9$\pm$0.8 & 35.1$\pm$0.2 & \textbf{39.1$\pm$0.2} \\ 
\bottomrule
}

% \setlength{\tabcolsep}{5pt}
% \ctable[
% % 	captionskip = -5pt,
%     pos=t,
%     caption = {Classification accuracy(\%) of various backbone network structures on CIFAR-100 with asymmetric noise 40\%.},
%     label = tab:modelcomparison,
%     doinside = \footnotesize
% ]{l|cccc}{
% }{
% \toprule
%  & DenseNet & VGG-19 & ResNet50 & MobileNetV2 \\ 
% \midrule
% Default              & 45.3$\pm$1.4                & 35.4$\pm$1.9  & 29.8$\pm$0.2    & 32.9$\pm$0.6       \\
% ActiveBias           & 50.3$\pm$0.6                & 31.4$\pm$0.7  & 28.2$\pm$0.5    & 38.1$\pm$0.6       \\
% Coteaching           & 47.3$\pm$1.4                & 35.6$\pm$0.5  & 32.6$\pm$0.2    & 31.9$\pm$0.8       \\
% SELFIE               & 52.8$\pm$0.5                & 39.3$\pm$0.3  & 32.8$\pm$0.2    & 35.1$\pm$0.2       \\
% \prgname{}           & \textbf{59.5$\pm$0.9} & \textbf{43.1$\pm$0.6}  & \textbf{35.9$\pm$0.4} & \textbf{39.1$\pm$0.2}  \\    
% \bottomrule
% }

%% Mixed noise
\noindent\emph{\textbf{Mixed noise.}}
%displays the test accuracies of DenseNet trained on datasets with mixed noise in which asymmetric noise and symmetric noise are added with different rates.
According to \tablename~\ref{tab:all}, the performance of \prgname{} achieves the best performance for mixed noise. As symmetric noise increases under the same noise level, the accuracy of \textit{Default} increases, presumably resulting from that symmetric noise is easy to distinguish. This result is in good agreement with results from Section \ref{loss_var_analysis}. We can also observe that the difference between \textit{Default} and other baselines reduces with more symmetric noise, indicating that symmetric noise does not require developed algorithms and lacks significance. Furthermore, \prgname{} can identify noise with high precision than other methods as indicated in \figurename~\ref{fig:precision}. These results of mixed noise imply our model's advantages against noisy real-world data, where symmetric and asymmetric noise may coexist.

%% Nearest Noise
\noindent\emph{\textbf{Nearest noise.}}
We can observe from \tablename~\ref{tab:all} that \prgname{} yields higher accuracy for nearest noise compared to other methods. The label precision of our method also surpasses other methods and continues to increase as training proceeds, while other methods appear to converge towards the end of training as shown in \figurename~\ref{fig:precision}.

%% High level noise
\noindent\emph{\textbf{High level noise.}}
To validate our method for another challenging problem, we conducted experiments on CIFAR-100 with larger noise rates for mixed noise and nearest noise. As shown in \tablename~\ref{tab:largenoise}, \prgname{} outperforms other state-of-the-art methods for larger noise rates. These results confirm that our method effectively downplays noisy samples and emphasizes clean and informative samples for all noise types.

%% Symmetric noise
\noindent\emph{\textbf{Symmetric noise.}} When adding symmetric noise, the true label can be included or excluded from the candidates of labels to be swapped, so we evaluated our method for both cases. We present the results of both definitions of symmetric noise in Appendix C due to lack of space and show that \prgname{} achieves comparable or better performance than other state-of-the-art methods for symmetric noise. 

%% Various networks
\noindent\emph{\textbf{Model architectures.}}
%\noindent\emph{\textbf{Network structures.}}
We evaluated whether \prgname{} is generic by comparing the performance of each method using various model architectures trained on CIFAR-100 with 40\% asymmetric noise. As shown in \tablename~\ref{tab:modelcomparison}, \prgname{} obtains the highest accuracy while producing consistent results despite changes in architectures. DenseNet (L=25, k=12) had the smallest architecture, thereby yielding better performance than those of other models such as ResNet50, which suffered severe overfitting. These results suggest that \prgname{} can be reliably applied to different model architectures.

\begin{figure}[t]
\centering
\begin{minipage}[t]{.48\textwidth}
\centering  
    \includegraphics[width=\textwidth]{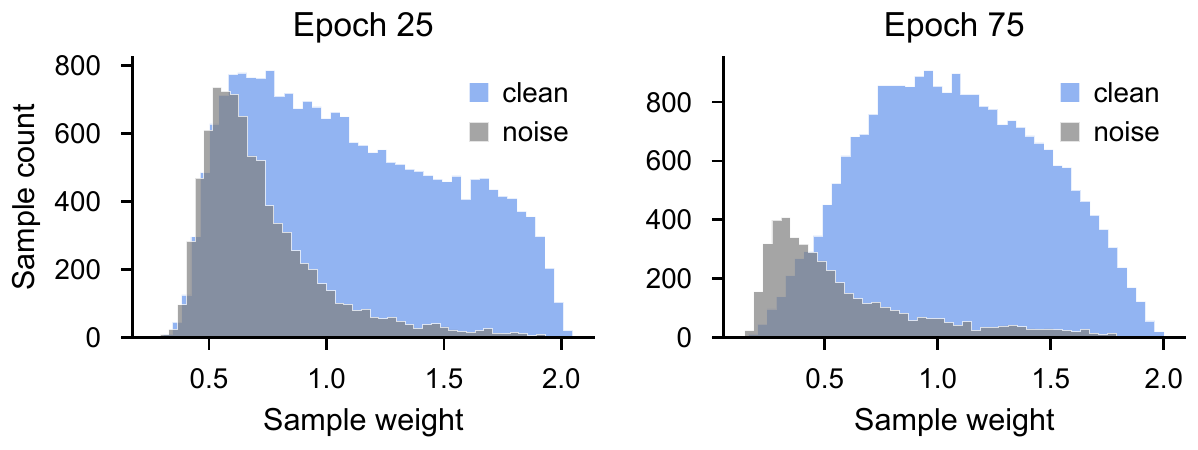}
    \subcaption{CIFAR-10}
    \label{fig:sample-weights-10}
\end{minipage}%
\hspace{10pt}
\begin{minipage}[t]{.48\textwidth}
\centering
    \includegraphics[width=\textwidth]{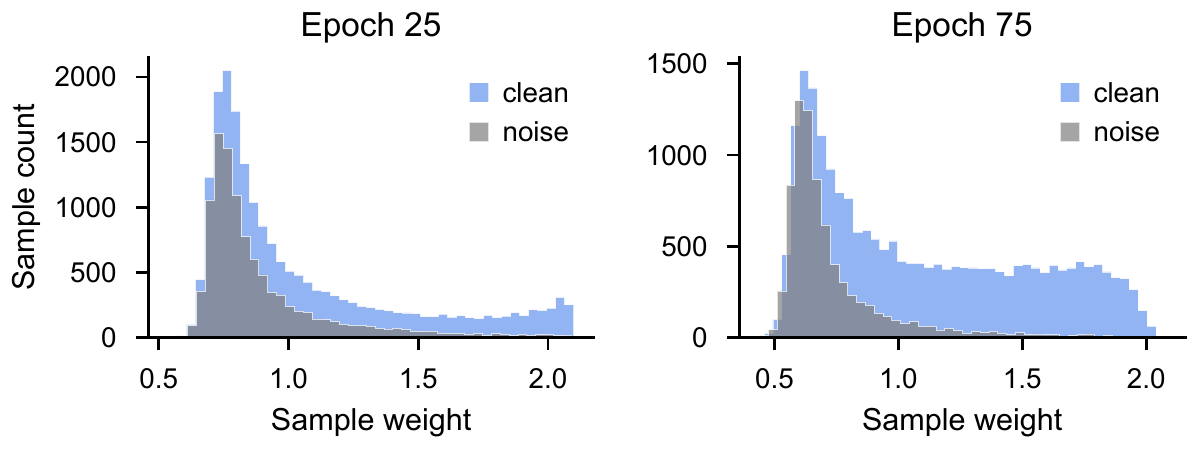}
    \subcaption{CIFAR-100}
    \label{fig:sample-weights-100}
\end{minipage}%

\caption{Changes of weight distribution on clean and noisy samples with non-zero weights. }
\label{fig:sample-weights}
\end{figure}

% \begin{figure}[t]
%     \centering
%     \hspace{-1em}
%     \subfloat[CIFAR-10]{
%         \label{fig:sample-weights-10}
%         \includegraphics[width=0.48\textwidth]{figures/5_sample_weights_cifar-10_40_asymmetric.pdf}}
%     % \qquad
%     %
%     \hspace{1.5pt}
%     \subfloat[CIFAR-100]{
%         \label{fig:sample-weights-100}
%         \includegraphics[width=0.48\textwidth]{figures/5_sample_weights_cifar-100_40_asymmetric.pdf}}
%     \vspace{-1.5pt}
%     \caption{Changes of weight distribution on clean and noisy samples with non-zero weights.}
%         \label{fig:sample-weights}

% \end{figure}

\setlength{\tabcolsep}{12pt}
\begin{table}[t]
\hspace{1em}
\begin{minipage}[t]{.45\linewidth}
    \centering
    \caption{ Results of variations in sample weighting on CIFAR-100 with 40\% asymmetric noise.  }
    \label{tab:ablation}
    \medskip
    {\small %  
    \begin{tabular}{l|cc}
        \toprule
        Variations & Acc.[\%] \\
        \midrule
        None (all the same)                & 57.1$\pm$1.0 \\ 
        Low loss            & 52.4$\pm$1.0 \\
        High uncertainty       & 55.7$\pm$1.7  \\ \hdashline
        Low loss, high uncertainty               & \textbf{59.5$\pm$0.9} \\
        \bottomrule
        %  removal \\ variation & keep samples (low var, $\hat{y} \neq y$)  & 56.5$\pm$0.6 \\
    \end{tabular}
    }% 
\end{minipage}%
\hspace{1.5em}
\begin{minipage}[t]{.45\linewidth}
    \centering
    \caption{ Results on Clothing1M. }
    \label{tab:clothing1m}
    {\small %  
        \begin{tabular}{l|cc}
        \toprule
        Methods                & Acc.[\%] \\
        \midrule
        % GCE  \citep{zhang2018generalized}   & 69.09         \\
        Forward loss  \cite{patrini2017making}         & 69.84         \\
        % M-DYR \citep{arazo2019unsupervised}              & 71.00            \\
        LCCN  \cite{yao2019safeguarded}                 & 71.63         \\
        Joint Optim.  \cite{tanaka2018joint}        & 72.16         \\
        DMI  \cite{xu2019l_dmi}                  & 72.46         \\
        MLNT  \cite{li2019learning}                 & 73.47         \\
        PENCIL  \cite{yi2019probabilistic}               & 73.49         \\ % \hdashline
        \textbf{\prgname{}} & \textbf{73.78}         \\ 
        \bottomrule
        \end{tabular}
    }%
\end{minipage}
\hspace{3.2em}
\end{table}

\subsection{ Empirical Analysis of Algorithm }
To comprehensively understand loss and uncertainty in our method, we conducted experiments on CIFAR-100 with 40\% asymmetric noise. \figurename~\ref{fig:sample-weights} displays how sample weights of \prgname{} change from soon after the warm-up stage (epoch 25) to convergence (epoch 75). 
We can observe that clean samples are allocated with larger weights, while noisy samples are allocated with smaller weights as training progresses. Moreover, the number of clean samples with non-zero weights increases, and the number of noisy samples, on the contrary, decreases (see also \figurename~\ref{fig:precision}). These results demonstrate our approach's effectiveness towards minimizing the impact of label noise and boosting the benefits of informative clean samples.

We also evaluated the weighting module for three approaches: placing larger weights on low loss, on high uncertainty, and treating every sample equally. As shown in \tablename~\ref{tab:ablation}, our weighting method outperforms other cases. Interestingly, giving equal weights achieved the best accuracy out of the three alternatives, while emphasizing samples with low loss led to the lowest accuracy. These results are parallel to \figurename~\ref{fig:asymmetric_noise_rate} in that \textit{Coteaching}, which focuses on low loss samples, performs worse than \textit{Active Bias}, which emphasizes high uncertainty samples. The detailed results of the ablation study is in Appendix D.

\subsection{Experiments on Clothing1M}
To further demonstrate our method’s effectiveness to realistic noise, we test on Clothing1M~\cite{xiao2015learning}, which comprises clothing data crawled from online shopping websites. Clothing1M consists of 1M images with real-world noisy labels and additional 50K, 14K, 10K verified clean data for training, validation and testing respectively. We retrain ResNet50 pretrained on ImageNet for 20 epochs using the 1M noisy dataset without any clean data in the training process. We use SGD with momentum of 0.9, $\epsilon=0.1$, $q=5$, $\gamma=5$, and $\tau=0.4$ because the estimated noise rate is 38\%~\cite{huang2019o2u, yi2019probabilistic}. The initial learning rate is 0.002 and is decreased by 10 every 5 epochs. For preprocessing, we resize images to 256x256 and randomly crop 224x224 from the resized images. This dataset is greatly imbalanced so we randomly select a relatively balanced subset of up to 35,000 samples for each class. 

As shown in Table~\ref{tab:clothing1m}, our method achieves 73.8\% accuracy, which is higher than recent state-of-the-art methods. For fair comparison, we do not include methods using different backbone models or any clean data during training.

\section{Conclusion}
In this paper, we have investigated the behavior of loss and uncertainty of samples for various noise. We have shown that for symmetric noise, noisy samples can be clearly identified with respect to either loss or uncertainty. For asymmetric noise\textemdash a more complex noisy label scenario that commonly occurs in real-world datasets\textemdash it is observed that considering both loss and uncertainty is necessary. Inspired by the findings, we have designed a novel method that aims at downplaying noisy samples while emphasizing clean and informative samples. Through series of experiments, we have demonstrated the effectiveness of \prgname{} when training with realistic synthetic label noise and real-world datasets as well as its generalizability in that it can be applied to any model architecture.

% \input{10.broader_impact.tex}

% \input{10.conclusion.tex}

% Acknowledgments of funding or assistance should be omitted for blind review

\bibliographystyle{unsrt}
\bibliography{reference}

\begin{thebibliography}{10}

\bibitem{deng2009imagenet}
Jia Deng, Wei Dong, Richard Socher, Li-Jia Li, Kai Li, and Li~Fei-Fei.
\newblock Imagenet: A large-scale hierarchical image database.
\newblock In {\em IEEE Conference on Computer Vision and Pattern Recognition},
  pages 248--255, 2009.

\bibitem{frenay2013classification}
Beno{\^\i}t Fr{\'e}nay and Michel Verleysen.
\newblock Classification in the presence of label noise: a survey.
\newblock {\em IEEE Transactions on Neural Networks and Learning Systems},
  25(5):845--869, 2013.

\bibitem{zhang2016understanding}
Chiyuan Zhang, Samy Bengio, Moritz Hardt, Benjamin Recht, and Oriol Vinyals.
\newblock Understanding deep learning requires rethinking generalization.
\newblock In {\em International Conference on Learning Representations}, 2017.

\bibitem{chang2017active}
Haw-Shiuan Chang, Erik Learned-Miller, and Andrew McCallum.
\newblock Active bias: Training more accurate neural networks by emphasizing
  high variance samples.
\newblock In {\em Advances in Neural Information Processing Systems}, pages
  1002--1012, 2017.

\bibitem{malach2017decoupling}
Eran Malach and Shai Shalev-Shwartz.
\newblock Decoupling ``when to update" from ``how to update".
\newblock In {\em Advances in Neural Information Processing Systems}, pages
  960--970, 2017.

\bibitem{schein2007active}
Andrew~I Schein and Lyle~H Ungar.
\newblock Active learning for logistic regression: an evaluation.
\newblock {\em Machine Learning}, 68(3):235--265, 2007.

\bibitem{joshi2009multi}
Ajay~J Joshi, Fatih Porikli, and Nikolaos Papanikolopoulos.
\newblock Multi-class active learning for image classification.
\newblock In {\em IEEE Conference on Computer Vision and Pattern Recognition},
  pages 2372--2379, 2009.

\bibitem{song2019selfie}
Hwanjun Song, Minseok Kim, and Jae-Gil Lee.
\newblock Selfie: Refurbishing unclean samples for robust deep learning.
\newblock In {\em International Conference on Machine Learning}, pages
  5907--5915, 2019.

\bibitem{hinton2007recognize}
Geoffrey~E Hinton.
\newblock To recognize shapes, first learn to generate images.
\newblock {\em Progress in Brain Research}, 165:535--547, 2007.

\bibitem{loshchilov2015online}
Ilya Loshchilov and Frank Hutter.
\newblock Online batch selection for faster training of neural networks.
\newblock In {\em International Conference on Learning Representations}, 2016.

\bibitem{malisiewicz2011ensemble}
Tomasz Malisiewicz, Abhinav Gupta, and Alexei Efros.
\newblock Ensemble of exemplar-svms for object detection and beyond.
\newblock In {\em IEEE International Conference on Computer Vision}, pages
  89--96, 2011.

\bibitem{shrivastava2016training}
Abhinav Shrivastava, Abhinav Gupta, and Ross Girshick.
\newblock Training region-based object detectors with online hard example
  mining.
\newblock In {\em IEEE Conference on Computer Vision and Pattern Recognition},
  pages 761--769, 2016.

\bibitem{han2018co}
Bo~Han, Quanming Yao, Xingrui Yu, Gang Niu, Miao Xu, Weihua Hu, Ivor Tsang, and
  Masashi Sugiyama.
\newblock Co-teaching: Robust training of deep neural networks with extremely
  noisy labels.
\newblock In {\em Advances in Neural Information Processing Systems}, pages
  8527--8537, 2018.

\bibitem{huang2019o2u}
Jinchi Huang, Lie Qu, Rongfei Jia, and Binqiang Zhao.
\newblock O2u-net: A simple noisy label detection approach for deep neural
  networks.
\newblock In {\em IEEE International Conference on Computer Vision}, pages
  3326--3334, 2019.

\bibitem{chen2019understanding}
Pengfei Chen, Ben~Ben Liao, Guangyong Chen, and Shengyu Zhang.
\newblock Understanding and utilizing deep neural networks trained with noisy
  labels.
\newblock In {\em International Conference on Machine Learning}, pages
  1062--1070, 2019.

\bibitem{goldberger2016training}
Jacob Goldberger and Ehud Ben-Reuven.
\newblock Training deep neural-networks using a noise adaptation layer.
\newblock In {\em International Conference on Learning Representations}, 2017.

\bibitem{reed2014training}
Scott Reed, Honglak Lee, Dragomir Anguelov, Christian Szegedy, Dumitru Erhan,
  and Andrew Rabinovich.
\newblock Training deep neural networks on noisy labels with bootstrapping.
\newblock In {\em International Conference on Learning Representations}, 2015.

\bibitem{van2015learning}
Brendan van Rooyen, Aditya Menon, and Robert~C Williamson.
\newblock Learning with symmetric label noise: The importance of being
  unhinged.
\newblock In {\em Advances in Neural Information Processing Systems}, pages
  10--18, 2015.

\bibitem{jiang2017mentornet}
Lu~Jiang, Zhengyuan Zhou, Thomas Leung, Li-Jia Li, and Li~Fei-Fei.
\newblock Mentornet: Learning data-driven curriculum for very deep neural
  networks on corrupted labels.
\newblock In {\em International Conference on Machine Learning}, pages
  2309--2318, 2018.

\bibitem{wang2018iterative}
Yisen Wang, Weiyang Liu, Xingjun Ma, James Bailey, Hongyuan Zha, Le~Song, and
  Shu-Tao Xia.
\newblock Iterative learning with open-set noisy labels.
\newblock In {\em IEEE Conference on Computer Vision and Pattern Recognition},
  pages 8688--8696, 2018.

\bibitem{konyushkova2017learning}
Ksenia Konyushkova, Raphael Sznitman, and Pascal Fua.
\newblock Learning active learning from data.
\newblock In {\em Advances in Neural Information Processing Systems}, pages
  4225--4235, 2017.

\bibitem{yang2018variance}
Yazhou Yang and Marco Loog.
\newblock A variance maximization criterion for active learning.
\newblock {\em Pattern Recognition}, 78:358--370, 2018.

\bibitem{rolnick2017deep}
David Rolnick, Andreas Veit, Serge Belongie, and Nir Shavit.
\newblock Deep learning is robust to massive label noise.
\newblock {\em arXiv preprint arXiv:1705.10694}, 2017.

\bibitem{ren2018learning}
Mengye Ren, Wenyuan Zeng, Bin Yang, and Raquel Urtasun.
\newblock Learning to reweight examples for robust deep learning.
\newblock In {\em International Conference on Machine Learning}, pages
  4334--4343, 2018.

\bibitem{yi2019probabilistic}
Kun Yi and Jianxin Wu.
\newblock Probabilistic end-to-end noise correction for learning with noisy
  labels.
\newblock In {\em IEEE Conference on Computer Vision and Pattern Recognition},
  pages 7017--7025, 2019.

\bibitem{arpit2017closer}
Devansh Arpit, Stanis{\l}aw Jastrz{\k{e}}bski, Nicolas Ballas, David Krueger,
  Emmanuel Bengio, Maxinder~S Kanwal, Tegan Maharaj, Asja Fischer, Aaron
  Courville, Yoshua Bengio, and Simon Lacoste-Julien.
\newblock A closer look at memorization in deep networks.
\newblock In {\em International Conference on Machine Learning}, pages
  233--242, 2017.

\bibitem{li2017learning}
Yuncheng Li, Jianchao Yang, Yale Song, Liangliang Cao, Jiebo Luo, and Li-Jia
  Li.
\newblock Learning from noisy labels with distillation.
\newblock In {\em IEEE International Conference on Computer Vision}, pages
  1910--1918, 2017.

\bibitem{liu2015classification}
Tongliang Liu and Dacheng Tao.
\newblock Classification with noisy labels by importance reweighting.
\newblock {\em IEEE Transactions on Pattern Analysis and Machine Intelligence},
  38(3):447--461, 2015.

\bibitem{xiao2015learning}
Tong Xiao, Tian Xia, Yi~Yang, Chang Huang, and Xiaogang Wang.
\newblock Learning from massive noisy labeled data for image classification.
\newblock In {\em IEEE Conference on Computer Vision and Pattern Recognition},
  pages 2691--2699, 2015.

\bibitem{seo2019combinatorial}
Paul~Hongsuck Seo, Geeho Kim, and Bohyung Han.
\newblock Combinatorial inference against label noise.
\newblock In {\em Advances in Neural Information Processing Systems}, pages
  1171--1181, 2019.

\bibitem{arazo2019unsupervised}
Eric Arazo, Diego Ortego, Paul Albert, Noel~E O'Connor, and Kevin McGuinness.
\newblock Unsupervised label noise modeling and loss correction.
\newblock In {\em International Conference on Machine Learning}, June 2019.

\bibitem{zhang2018mixup}
Hongyi Zhang, Moustapha Cisse, Yann~N Dauphin, and David Lopez-Paz.
\newblock mixup: Beyond empirical risk minimization.
\newblock In {\em International Conference on Learning Representations}, 2018.

\bibitem{huang2017densely}
Gao Huang, Zhuang Liu, Laurens Van Der~Maaten, and Kilian~Q Weinberger.
\newblock Densely connected convolutional networks.
\newblock In {\em IEEE Conference on Computer Vision and Pattern Recognition},
  pages 4700--4708, 2017.

\bibitem{kim2018nsml}
Hanjoo Kim, Minkyu Kim, Dongjoo Seo, Jinwoong Kim, Heungseok Park, Soeun Park,
  Hyunwoo Jo, KyungHyun Kim, Youngil Yang, Youngkwan Kim, et~al.
\newblock Nsml: Meet the mlaas platform with a real-world case study.
\newblock {\em arXiv preprint arXiv:1810.09957}, 2018.

\bibitem{sung2017nsml}
Nako Sung, Minkyu Kim, Hyunwoo Jo, Youngil Yang, Jingwoong Kim, Leonard Lausen,
  Youngkwan Kim, Gayoung Lee, Donghyun Kwak, Jung-Woo Ha, et~al.
\newblock Nsml: A machine learning platform that enables you to focus on your
  models.
\newblock {\em arXiv preprint arXiv:1712.05902}, 2017.

\bibitem{shu2019meta}
Jun Shu, Qi~Xie, Lixuan Yi, Qian Zhao, Sanping Zhou, Zongben Xu, and Deyu Meng.
\newblock Meta-weight-net: Learning an explicit mapping for sample weighting.
\newblock In {\em Advances in Neural Information Processing Systems}, pages
  1917--1928, 2019.

\bibitem{patrini2017making}
Giorgio Patrini, Alessandro Rozza, Aditya Krishna~Menon, Richard Nock, and
  Lizhen Qu.
\newblock Making deep neural networks robust to label noise: A loss correction
  approach.
\newblock In {\em IEEE Conference on Computer Vision and Pattern Recognition},
  pages 2233--2241, 2017.

\bibitem{yao2019safeguarded}
Jiangchao Yao, Hao Wu, Ya~Zhang, Ivor~W Tsang, and Jun Sun.
\newblock Safeguarded dynamic label regression for noisy supervision.
\newblock In {\em Proceedings of the AAAI Conference on Artificial
  Intelligence}, volume~33, pages 9103--9110, 2019.

\bibitem{tanaka2018joint}
Daiki Tanaka, Daiki Ikami, Toshihiko Yamasaki, and Kiyoharu Aizawa.
\newblock Joint optimization framework for learning with noisy labels.
\newblock In {\em IEEE Conference on Computer Vision and Pattern Recognition},
  pages 5552--5560, 2018.

\bibitem{xu2019l_dmi}
Yilun Xu, Peng Cao, Yuqing Kong, and Yizhou Wang.
\newblock $l_{DMI}$: A novel information-theoretic loss function for training
  deep nets robust to label noise.
\newblock In {\em Advances in Neural Information Processing Systems}, pages
  6222--6233, 2019.

\bibitem{li2019learning}
Junnan Li, Yongkang Wong, Qi~Zhao, and Mohan~S Kankanhalli.
\newblock Learning to learn from noisy labeled data.
\newblock In {\em IEEE Conference on Computer Vision and Pattern Recognition},
  pages 5051--5059, 2019.

\bibitem{ma2018dimensionality}
Xingjun Ma, Yisen Wang, Michael~E Houle, Shuo Zhou, Sarah~M Erfani, Shu-Tao
  Xia, Sudanthi Wijewickrema, and James Bailey.
\newblock Dimensionality-driven learning with noisy labels.
\newblock In {\em International Conference on Machine Learning}, 2018.

\end{thebibliography}

\begin{alphasection}
\newpage
\section{Types of Noise}

There are four types of noise used in this paper: asymmetric noise, symmetric noise, mixed noise, and nearest noise. The noise rate is denoted by $\tau$. \figurename~\ref{fig:noise_types} displays the noise transition matrices of each noise type.

As can be seen from \figurename~\ref{fig:pair}, asymmetric noise swaps labels between two classes with a probability of $\tau$. Asymmetric noise is problematic and similar to real-world noise~\cite{ren2018learning, yi2019probabilistic}. We, therefore, place more emphasis on the results of asymmetric noise to provide a promising method for realistic noise.

For symmetric noise, which is less practical than asymmetric noise~\cite{han2018co,ren2018learning}, the true label can be swapped to any other label. There are two definitions of symmetric noise in prior works. As shown in \figurename~\ref{fig:symmetry}, one popular label noise criterion is random labeling while the true label is not selected ~\cite{jiang2017mentornet, wang2018iterative}. In this case, the probability of being swapped to another label is uniformly distributed with value $\frac{\tau}{k-1}$, so the sum of probabilities being swapped becomes the noise rate $\tau$. As displayed in \figurename~\ref{fig:symmetry2}, another criterion for symmetric noise addition consists of randomly
selecting labels for a percentage of the training data using
all possible labels (i.e. the true label could be randomly
maintained)~\cite{zhang2018mixup, tanaka2018joint}. The probability of being swapped to another label is thus $\frac{\tau}{k}$.

Mixed noise is when asymmetric noise and symmetric noise are added together as can be seen in \figurename~\ref{fig:mixed}. Mixed noise represents the scenario of noise mainly injected from another class along with some random noise. We have experimented with mixed noise to produce more realistic noise.

Nearest noise is used to simulate confusions between visually similar classes~\cite{seo2019combinatorial}. As shown in \figurename~\ref{fig:nearest}, the probabilities of being swapped are different for each class. For the nearest neighbor search, we use a confusion matrix of a pretrained network of the dataset. The validation accuracy of the pretrained network trained on CIFAR-100 was 53.12\%.

\begin{figure}[!htb]
\centering
\begin{minipage}[t]{.25\textwidth}
\centering  
    \includegraphics[width=\textwidth]{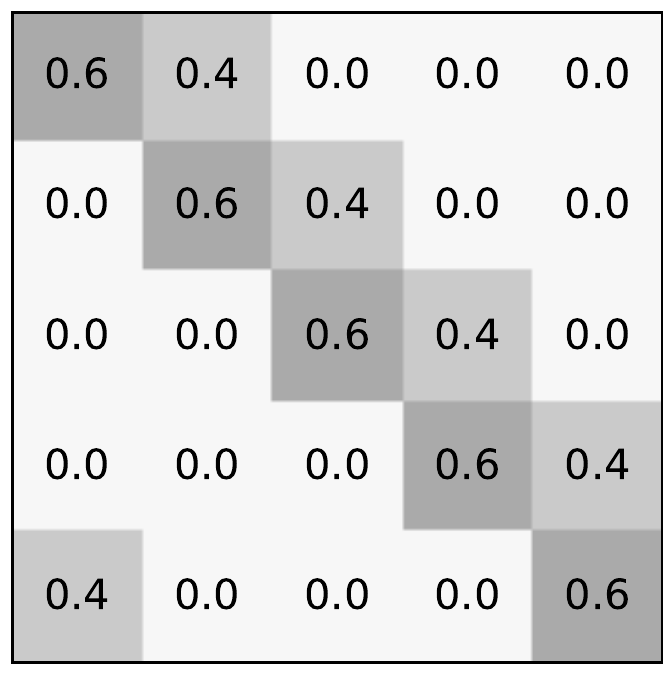}
    \subcaption{Asymmetric noise.}
    \label{fig:pair}
\end{minipage}%
\hspace{10pt}
\begin{minipage}[t]{.25\textwidth}
\centering
    \includegraphics[width=\textwidth]{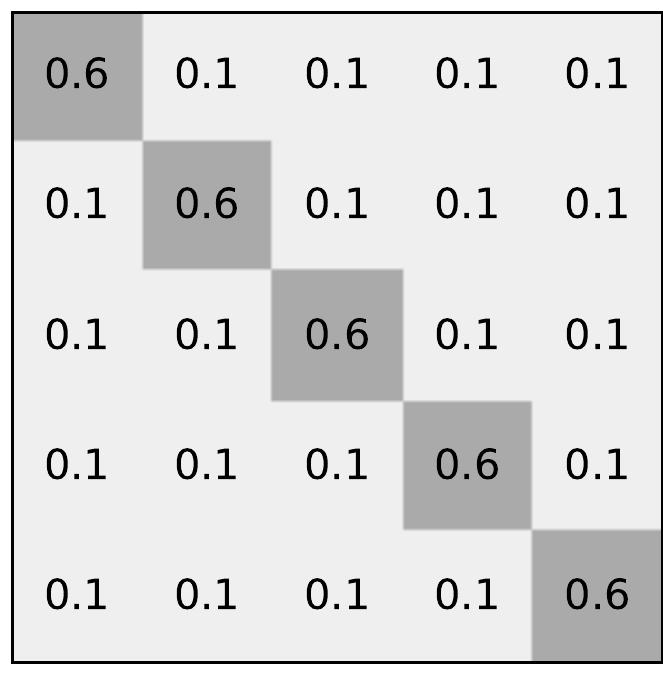}
    \subcaption{Symmetric noise excluding true labels.}
    \label{fig:symmetry}
\end{minipage}%
\hspace{10pt}
\begin{minipage}[t]{.25\textwidth}
\centering
    \includegraphics[width=\textwidth]{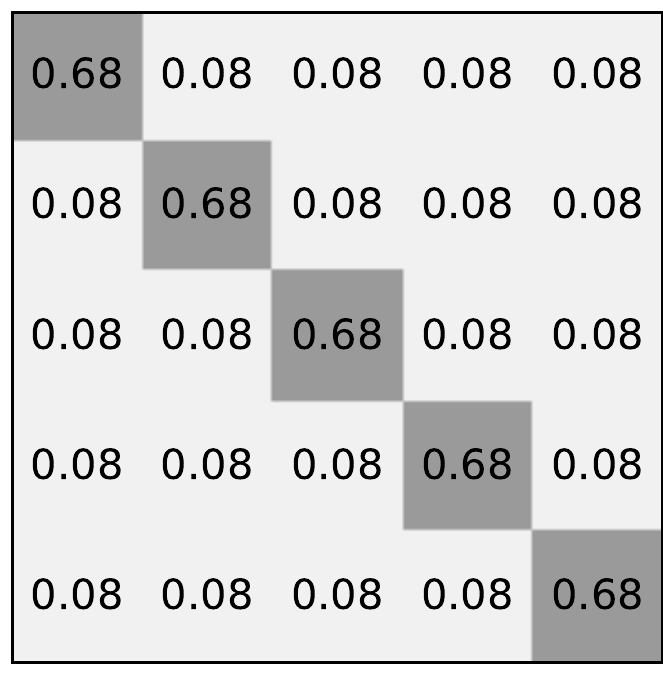}
    \subcaption{Symmetric noise including true labels.}
    \label{fig:symmetry2}
\end{minipage}%
\vspace{10pt}
\begin{minipage}{.25\textwidth}
\centering  
    \includegraphics[width=\textwidth]{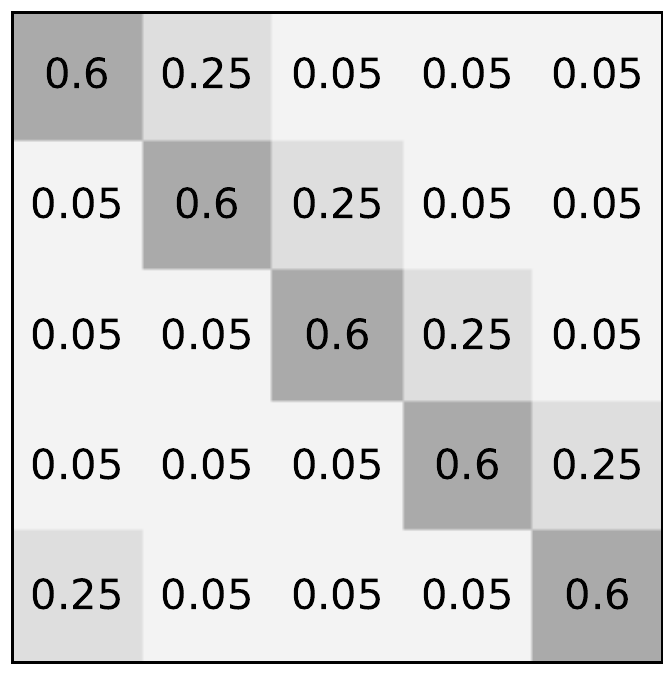}
    \subcaption{Mixed noise.}
    \label{fig:mixed}
\end{minipage}
\hspace{10pt}
\begin{minipage}{.25\textwidth}
\centering  
    \includegraphics[width=\textwidth]{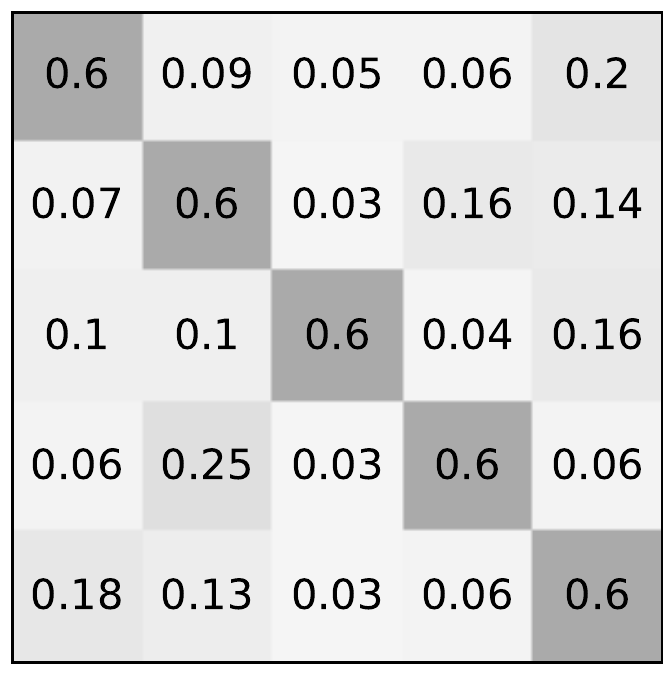}
    \subcaption{Nearest noise.}
    \label{fig:nearest}
\end{minipage}

\caption{Types of 40\% noise transition matrices for 5 classes. Mixed noise consists of 20\% asymmetric noise and 20\% symmetric noise.  }
\label{fig:noise_types}
\end{figure}

\newpage % a
\section{Loss and Uncertainty During Training }

\noindent \figurename~\ref{fig:symmetry_noise_rate} and~\ref{fig:pair_noise_rate} show how loss and uncertainty change throughout the training process. Clean (60\%) and noisy (40\%) samples are divided vertically. The green and red colors represent low and high loss samples, and the solid and stripe patterns represent low and high uncertainty samples, respectively. 

As can be seen from the figures, after 10 epochs, there are subtle changes in proportions concerning loss and uncertainty during training. For symmetric noise, the proportions remain constant throughout the training process (see \figurename~\ref{fig:symmetry_noise_rate}). For asymmetric noise, although there are slight changes throughout training, the changes are negligible (see \figurename~\ref{fig:pair_noise_rate}). We can also observe that asymmetric noise is more problematic than symmetric noise, because samples for each combination are distributed fairly evenly for noisy and clean samples.

\begin{figure*}[!htb]
    \centering
    \includegraphics[width=.9\textwidth]{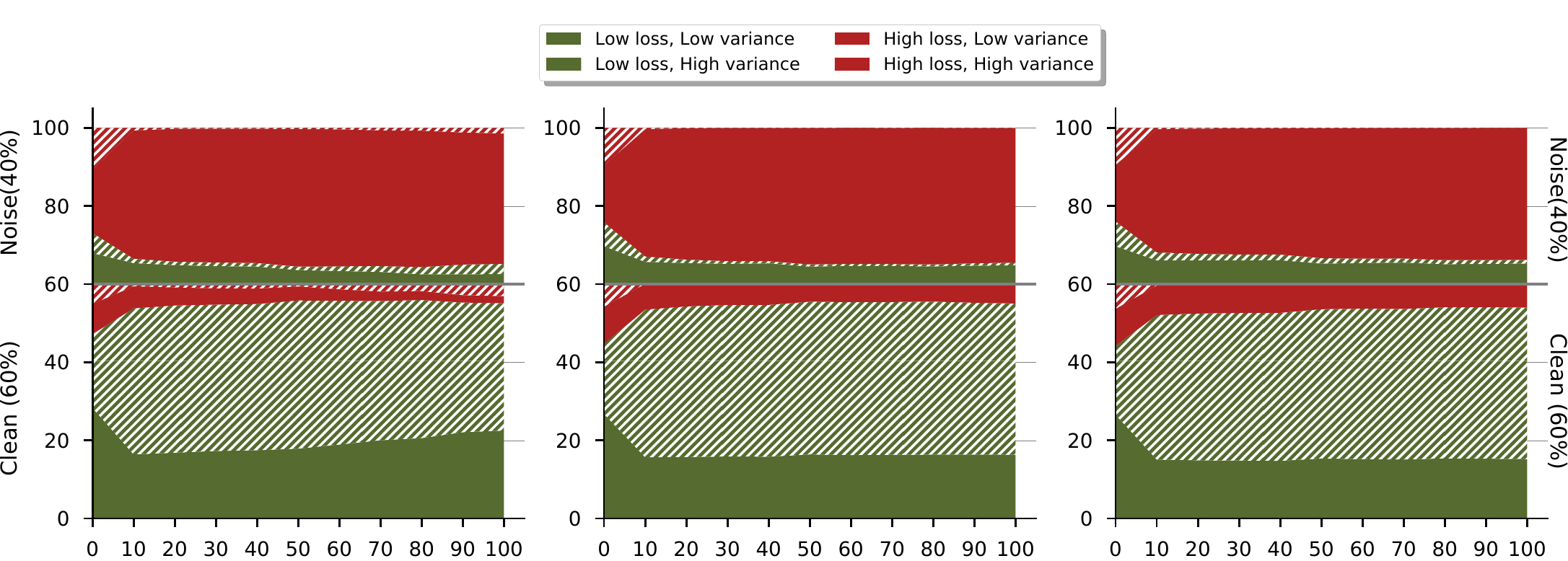}
    \begin{minipage}[t]{.33\linewidth}
        \vspace{-7pt}
        \centering
        \subcaption{CIFAR-10.}\label{symmetry_cifar10}
    \end{minipage}%
    \begin{minipage}[t]{.28\linewidth}
        \vspace{-7pt}
        \centering
        \subcaption{CIFAR-100.}\label{symmetry_cifar100}
    \end{minipage}
    \begin{minipage}[t]{.28\linewidth}
        \vspace{-7pt}
        \centering
        \subcaption{Tiny ImageNet.}\label{symmetry_tinyimagenet}
    \end{minipage}
    \vspace{-5pt}
    \caption{Loss and uncertainty of datasets with 40\% \textbf{symmetric noise} during training. } \label{fig:symmetry_noise_rate}
    \vspace{-10pt}
\end{figure*}

\begin{figure*}[!htb]
    \centering
    \includegraphics[width=.9\textwidth]{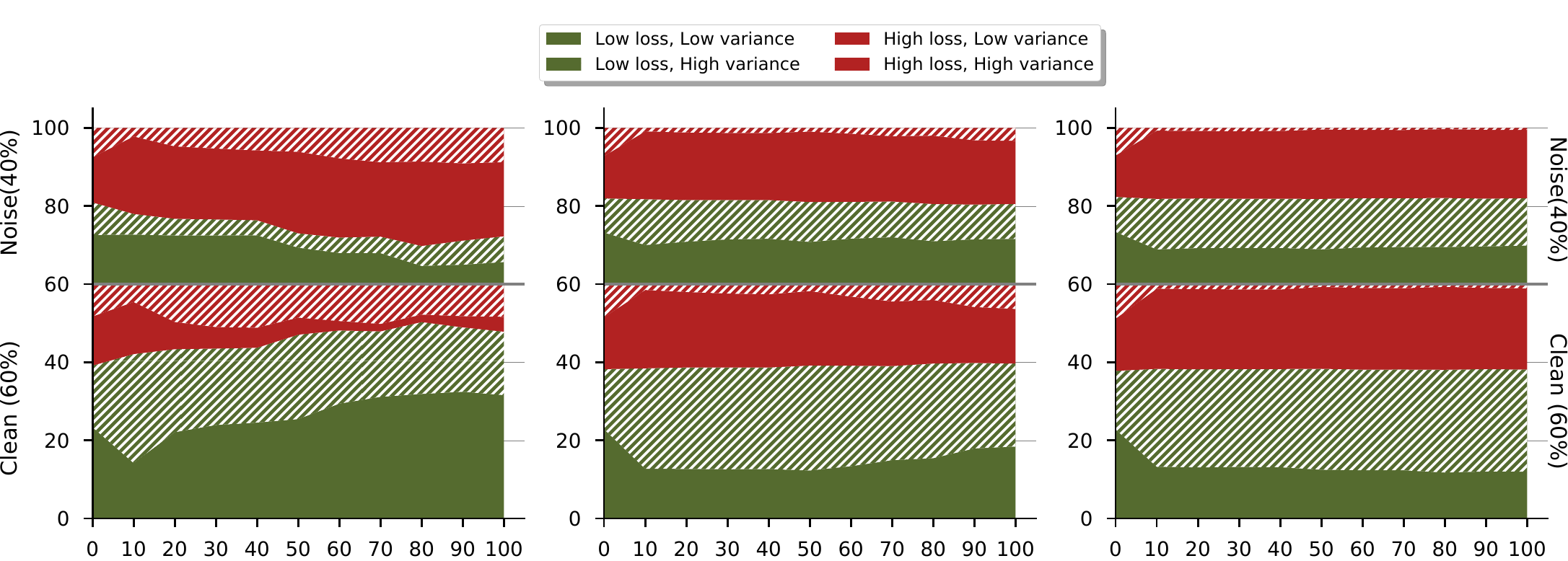}
    \begin{minipage}[t]{.33\linewidth}
        \vspace{-7pt}
        \centering
        \subcaption{CIFAR-10.}\label{pair_cifar10}
    \end{minipage}%
    \begin{minipage}[t]{.28\linewidth}
        \vspace{-7pt}
        \centering
        \subcaption{CIFAR-100.}\label{pair_cifar100}
    \end{minipage}
    \begin{minipage}[t]{.28\linewidth}
        \vspace{-7pt}
        \centering
        \subcaption{Tiny ImageNet.}\label{pair_tinyimagenet}
    \end{minipage}
    \vspace{-5pt}
    \caption{Loss and uncertainty of datasets with 40\% \textbf{asymmetric noise} during training. } \label{fig:pair_noise_rate}
\end{figure*}

% \clearpage
 % b
% \twocolumn
\section{Symmetric Noise Results}

More research has focused on symmetric noise or random labeling, so we compare our results with recent state-of-the-art methods: bootstrapping, forward loss, mixup, MentorNet, D2L, and MD-DYR-SH, which uses dynamic mixup, soft to hard dynamic bootstrapping with regularization. As explained in Appendix A, when adding symmetric noise, the true label can be included or excluded from the candidates of labels to be swapped. We have hence evaluated our method for both cases. 

\tablename~\ref{tab:random-labeling} displays the classification accuracy results of 40\% symmetric noise for CIFAR-10 and CIFAR-100. \prgname{} was trained on datasets using PreAct ResNet-18 following the experimental settings of Arazo et al.~\cite{arazo2019unsupervised}. We report the results from the paper and as shown from the table, our method achieves accuracy higher than or equivalent to other state-of-the-art methods for both symmetric noise types. However, the results on symmetric noise excluding true labels should be interpreted with care because some methods employed different architectures and used clean data during training such as in~\cite{jiang2017mentornet}.

\begin{table}[t]
\caption{Classification accuracy on benchmark datasets with symmetric noise using all labels (top) and excluding true labels (bottom). Wide ResNet is denoted by WRN, Generic CNN is GCNN, ResNet is RN, and PreAct ResNet is PRN.  } \label{tab:random-labeling}
\begin{subtable}{\textwidth}
%   \caption{Random labeling with true labels}\label{tab:sym-with-true}
    \centering
    \small{
  \begin{tabular}{l|ccccc}
      \toprule
    & Bootstrapping~\cite{reed2014training} & Forward loss~\cite{patrini2017making} & mixup~\cite{zhang2018mixup} & MD-DYR-SH~\cite{arazo2019unsupervised} & \prgname{} \\
     \midrule
    \midrule
    CIFAR-10  & 86.8 & 86.8         & \textbf{95.6}      & 93.8     & 94.1                        \\
    CIFAR-100 & 62.1 & 61.5         & 67.8     & 73.9     & \textbf{74.2} \\
    \bottomrule
  \end{tabular}
  }
\end{subtable}%
\vspace{0.5em}
\begin{subtable}{\textwidth} % 
    % \caption{Random labeling without true labels}\label{tab:sym-wo-true}
    \centering
    \small{
  \begin{tabular}{l|cccc}
    \toprule
      & MentorNet~\cite{jiang2017mentornet} &  D2L~\cite{ma2018dimensionality} & MD-DYR-SH~\cite{arazo2019unsupervised} & \prgname{} \\
     \midrule
     Architecture  & WRN-101           & GCNN-12/RN-44   & PRN-18    & PRN-18                      \\
     \midrule
     \midrule
    CIFAR-10  & 92.0                & 85.1         & 93.8      & \textbf{94.0}                          \\
    CIFAR-100 & 73.0                & 62.2         & 73.7      & \textbf{74.0}          \\
    \bottomrule

  \end{tabular}
  }
\end{subtable}

\end{table}
 % c

\section{Ablation Study on Hyperparameters}

 We present an ablation study on hyperparameters: $q$ (queue size), $\gamma$ (warm-up), and $\epsilon$ (threshold). We recorded the classification accuracies (\%) over three runs for CIFAR-100 with 40\% asymmetric noise (default q=15, g=25, e=0.1). As shown in \tablename~\ref{tab:hyperparam}, the results do not greatly depend on hyperparameters except for the strictest case (e=0). Therefore, it can be concluded that our method is practical due to its insensitivity to hyperparameters.

\setlength{\tabcolsep}{14pt}
\setlength{\extrarowheight}{.1em}
\ctable[
    pos=h,
    % captionsleft,
    caption = {Ablation study on hyperparameters: $q,\gamma, \epsilon$.},
    center,
    label = tab:hyperparam,
    doinside = \footnotesize
]{ccc|c}{
}{
\toprule
$q$  & $\gamma$ & $\epsilon$ & Accuracy (\%)           \\
\midrule
15               & 25                & 0.1                   & 59.5 $\pm$ 0.9 \\ \hline
5                & 25                & 0.1                   & 58.1 $\pm$ 0.8 \\
25               & 25                & 0.1                   & 57.6 $\pm$ 1.2 \\ \hline
15               & 15                & 0.1                   & 57.9 $\pm$ 0.9 \\
15               & 35                & 0.1                   & 58.8 $\pm$ 0.3 \\
15               & 45                & 0.1                   & 58.6 $\pm$ 0.2 \\ \hline
15               & 25                & 0.0                   & 55.9 $\pm$ 0.1 \\
15               & 25                & 0.2                   & 59.4 $\pm$ 0.6 \\
15               & 25                & 0.3                   & 59.3 $\pm$ 0.3 \\

\bottomrule
}

\newpage % d
\end{alphasection}

\end{document}